\documentclass[journal,twocolumn,final]{IEEEtran}
\usepackage{generic}
\usepackage{cite}
\usepackage{amsmath,amssymb,amsfonts}
\usepackage{mathtools}
\usepackage{algorithm}
\usepackage{algorithmic}
\usepackage{graphicx}
\usepackage{textcomp}
\let\labelindent\relax
\usepackage{enumitem}
\usepackage{threeparttable}
\usepackage{kantlipsum, lipsum}
\usepackage{dsfont}
\usepackage{float}
\usepackage{subcaption}
\usepackage{multirow}
\usepackage{makecell}
\usepackage{tabularx}
\usepackage{booktabs}
\usepackage{hyperref}
\usepackage{cleveref}

\usepackage{amsthm}
\Crefname{table}{Table}{Tables}
\Crefname{figure}{Fig.}{Fig.}
\usepackage[normalem]{ulem}
\usepackage{colortbl}
\usepackage{orcidlink}

\newtheorem{theorem}{Theorem} %[section]
\newtheorem{lemma}[theorem]{Lemma}
\newtheorem{proposition}{Proposition} %[section]

\newtheorem{definition}{Definition} %[section]
\begin{document}
\title{Towards Intrinsically Calibrated Uncertainty Quantification in Industrial Data-Driven Models via Diffusion Sampler}

\author{
  Yiran Ma~\orcidlink{0009-0004-6749-5121}, Jerome {Le Ny}~\orcidlink{0000-0002-3417-4135}, \IEEEmembership{Senior Member, IEEE}, Zhichao Chen~\orcidlink{0000-0001-5785-0741}, and Zhihuan Song~\orcidlink{0000-0003-4098-6479}
  % Yiran Ma, Jerome {Le Ny}, \IEEEmembership{Senior Member, IEEE}, Zhichao Chen, and Zhihuan Song
  \thanks{This manuscript has been accepted for publication in IEEE Transactions on Industrial Informatics.
  © IEEE. Reuse of this material is subject to IEEE copyright restrictions. See \url{https://www.ieee.org/publications/rights/index.html} for more information.}
  \thanks{This work was supported by the National Natural Science Foundation of China under Grant Number 62473103 and was also supported by the China Postdoctoral Science Foundation under Grant Number 2025M781449. Part of this work was done while the first author was visiting Polytechnique Montreal and GERAD. (Corresponding author: Zhihuan Song and Zhichao Chen)}
  \thanks{Yiran Ma is with the College of Control Science and Engineering, Zhejiang University, Hangzhou, 310027, China (e-mail: {mayiran@zju.edu.cn}).}
  \thanks{Jerome Le Ny is with the Electrical Engineering Department, Polytechnique Montreal, and with GERAD, Montreal, QC H3T 1J4, Canada (e-mail: jerome.le-ny@polymtl.ca).}
  \thanks{Zhichao Chen is with the State Key Lab of General AI, School of Intelligence Science and Technology, Peking University, Beijing 100871, China (e-mail: 12032042@zju.edu.cn).}
  \thanks{Zhihuan Song is with the Guangdong Provincial Key Laboratory of Petrochemical Equipment Fault Diagnosis, Guangdong University of Petrochemical Technology, Maoming, 525000, China, and also with the College of Control Science and Engineering, Zhejiang University, Hangzhou, 310027, China (e-mail: songzhihuan@zju.edu.cn).}
}

\maketitle

\begin{abstract}
  In modern process industries, data-driven models are important tools for real-time monitoring when key performance indicators are difficult to measure directly. While accurate predictions are essential, reliable uncertainty quantification (UQ) is equally critical for safety, reliability, and decision-making, but remains a major challenge in current data-driven approaches. In this work, we introduce a diffusion-based posterior sampling framework that inherently produces well-calibrated predictive uncertainty via faithful posterior sampling, eliminating the need for post-hoc calibration. In extensive evaluations on synthetic distributions, the Raman-based phenylacetic acid soft sensor benchmark, and a real ammonia synthesis case study, our method achieves practical improvements over existing UQ techniques in both uncertainty calibration and predictive accuracy. These results highlight diffusion samplers as a principled and scalable paradigm for advancing uncertainty-aware modeling in industrial applications.
\end{abstract}


\section{Introduction}\label{sec:intro}
Data-driven models have become essential tools in modern process industries, enabling the indirect estimation of key performance indicators that are difficult to physically measure in real time \cite{kadlec2009data}. However, in practice, such models often suffer from a lack of trust from industrial practitioners, which substantially limits their deployment in safety-critical and decision-driven scenarios. A major reason is that most existing data-driven models provide only point predictions without reliable measures of confidence, making it difficult to assess the risk associated with their outputs and to support robust operational decision making \cite{ma2024heat}.

Formally, let $x \in \mathbb{R}^d$ denote the vector of easy-to-measure variables and $y \in \mathbb{R}$ the target performance indicator. In most existing data-driven modeling practices, given a historical dataset $\mathcal{D}={(x_i, y_i)}{i=1}^N$, a deterministic mapping $\hat{y} = f_\theta(x), \ \theta \in \Theta$, is learned by minimizing the prediction error (e.g., in a mean-square sense). However, such point estimators provide no explicit mechanism to account for either data noise (e.g., measurement errors and process disturbances) or model uncertainty arising from limited data and model misspecification. Consequently, they often yield overconfident yet incorrect predictions, which is particularly dangerous in safety-critical and optimization-driven process control.

Therefore, beyond providing point predictions, it is crucial to assess their reliability. To meet this demand, this work focuses on uncertainty quantification, aiming to deliver not only accurate predictions of $y$ but also a calibrated predictive distribution $p(y \mid x, \mathcal{D})$. Such a distribution allows for constructing statistically meaningful credible intervals. For example, a nominal 95\% credible interval should empirically contain the true value approximately 95\% of the time. This capability is particularly valuable for industrial tasks such as reliable decision-making \cite{kochenderfer2015decision}, risk-sensitive control \cite{hansen2001robust}, and optimization \cite{sahinidis2004optimization}. Moreover, it underpins data-efficient learning by identifying areas of uncertainty in the model’s knowledge \cite{gal2017deep}.

Bayesian approaches naturally address this need because, when faced with multiple plausible and competing explanations for observed data, they take all possibilities into account rather than prematurely committing to any single explanation \cite{mackay1992practical, neal1996bayesian}.
This helps in understanding what a data-driven model does not know, which is important for safety-critical scenarios.
In practice, Bayesian models predicts by marginalization $p(y|x)=\int p(y|x,\theta)p(\theta|\mathcal{D})d\theta$, thus requiring a proper likelihood function $p(y|x,\theta)$ and the complete representation of the posterior $p(\theta|\mathcal{D})$.
Although conjugate Bayesian models underpinned many early successes in industrial data-driven modeling \cite{ge2018process}, typically by adopting linear and Gaussian likelihood, conjugate exponential-family likelihoods, or low-dimensional parameterizations that admit closed-form posteriors, these successes rely on strong structural assumptions. In the nonlinear, heteroscedastic noise and high-dimensional regimes of modern complex industrial practice, more complex models such as neural networks (NNs) are used \cite{sun2021survey}, where closed-form posteriors are no longer available; explicit Bayesian inference becomes either inapplicable or computationally prohibitive. Therefore, the problem of predictive uncertainty quantification (UQ) for industrial models has not been comprehensively addressed in the literature.

Furthermore, due to the intractability of exact Bayesian inference in modern nonlinear and high-dimensional industrial process models, most existing industrial UQ methods resort to classical approximate inference schemes or heuristic uncertainty estimators, thus requiring post-hoc calibration to compensate for systematic bias in uncertainty estimates. In industrial scenarios, however, obtaining ground-truth data often relies on time-consuming laboratory analysis; consequently, reserving a portion of this already scarce data or collecting a new independent set for post-hoc calibration \cite{liu2019understanding, d2021annealed, ma2024heat} is practically expensive, time-consuming, and sometimes even infeasible.
Consequently, the central challenge in designing industrial UQ frameworks lies in achieving intrinsically calibrated predictive uncertainty. Namely, uncertainty estimates that are reliable by construction and do not depend on additional post-hoc calibration or extra ground-truth data.

Fundamentally, the need for post-hoc calibration of existing methods reflects a deeper issue: the underlying approximate inference or sampling procedures fail to faithfully represent the true posterior distribution, leading to systematic under-coverage or mis-calibration. In complex industrial models, the posterior is often high-dimensional and non-Gaussian, making faithful posterior approximation particularly challenging.
Existing methods typically fail to provide reliable posterior coverage for three fundamentally different reasons.
First, methods based on restrictive parameterizations (e.g., Monte Carlo dropout \cite{yang2022remaining, cao2023parallel, gal2016dropout} and Mean-field variational inference (MFVI) \cite{lee2021uncertainty}) impose unimodal or factorized posterior families that cannot represent complex structures.
Second, sampling-based methods such as SG-MCMC \cite{zhang2023nonasymptotic} often suffer from prohibitively slow mixing in high-dimensional landscapes.
Third, particle-based methods such as SVGD \cite{ma2024heat} rely on finite-particle kernelized approximations that degenerate in high dimensions, leading to particle collapse or insufficient mode coverage \cite{liu2019understanding, d2021annealed, ma2024heat}.
As a result, these methods produce systematically under-covered posteriors, resulting in miscalibrated uncertainty estimates in practice and often necessitating post-hoc calibration on an additional independent dataset. Therefore, the key technical difficulty lies in ensuring faithful coverage of the entire posterior support, rather than approximating only the dominant mode(s).

To achieve a more faithful approximation of the posterior distribution, we adopt the Schrödinger bridge (SB) perspective \cite{leonard2013survey}. This theoretically grounded framework casts posterior inference as a stochastic transport problem: finding a path-space distribution whose terminal marginal matches the target posterior while minimizing the KL divergence from a reference diffusion process.
However, the main difficulty lies in reformulating this strict transport problem into a tractable, unconstrained objective to enable efficient end-to-end training via gradient descent.
While earlier approaches relied on iterative proportional fitting \cite{kullback1968probability}, recent work shows that the SB problem admits a stochastic optimal control (SOC) interpretation, offering a principled and well-posed objective for inference.
Therefore, recent advancements have increasingly focused on learning-based methods leveraging diffusion processes \cite{huang2021schrodinger, zhang2021path, vargas2023denoising}, which are named \emph{diffusion samplers}. These diffusion samplers are particularly compelling due to their scalability in high-dimensional settings. Recently, adjoint matching \cite{havens2025adjoint, liu2025adjoint} has further improved the efficiency of training diffusion samplers, making them more practical for complex models.
While these methods yield principled samplers from complex posterior or energy distributions that can be used to compute uncertainty measures, explicit UQ applications of diffusion samplers are still relatively limited; most existing work focuses on sampling methodology and theoretical analysis rather than dedicated UQ case studies.

In this work, we therefore build upon the SB formulation to construct a robust diffusion sampler for industrial data-driven modeling, aiming to provide calibrated uncertainty by capturing complex structure of posterior distributions inherent in such practical industrial models.

To summarize, the contributions of this paper are as follows:
\begin{enumerate}
  \item We introduce the diffusion sampler for posterior sampling, providing a principled alternative to conventional industrial data-driven modeling and establishing a theoretically grounded framework for UQ.
  \item We demonstrate the practical value of the framework on synthetic benchmarks and real-world industrial modeling tasks, showing improved uncertainty quality without the need for post-hoc calibration.
  \item Additional analyses show that the method exhibits robustness to hyperparameters and smooth optimization dynamics, highlighting diffusion samplers as a promising direction for uncertainty-aware industrial models.
\end{enumerate}

\section{Preliminaries}
\subsection{Uncertainty Quantification in Machine Learning}
Machine learning models, in particular deep neural networks, are often regarded as black-box predictors, producing outputs without revealing the confidence or reliability of their decisions. In many real-world scenarios, such as the industrial applications we study, however, relying solely on point predictions is insufficient, as overconfident yet incorrect predictions can lead to costly or unsafe outcomes. This has motivated a growing demand for UQ, which aims to complement point predictions with principled measures of uncertainty.

The central objective in UQ is \emph{calibration}, which formalizes the statistical consistency between predicted probabilities and empirical frequencies. A well-calibrated model satisfies \Cref{def:cali}.

\begin{definition}[Calibration] \label{def:cali}
  Given an input $x$, consider a probabilistic model that provides a predictive distribution $p(y \mid x)$ over the target $y$.
  Let $\mathcal{C}_{1-\alpha}(x)$ denote the $(1-\alpha)$-credible set induced by $p(y \mid x)$, representing the region that contains true $y$ with a nominal coverage level of $1-\alpha$.
  The model is perfectly calibrated if, for any coverage level $1-\alpha \in (0,1)$, the following holds:
  \begin{equation}
    \Pr\big(y \in \mathcal{C}_{1-\alpha}(x)\big) = 1-\alpha. \label{eq:cali}
  \end{equation}
\end{definition}

Intuitively, (\ref{eq:cali}) means that the probability of observations falling inside the empirical $(1-\alpha)$-credible interval matches the expected coverage $1-\alpha$. Based on this understanding, calibration for regression tasks can be defined, see \Cref{def:calireg}.

\begin{definition}[Calibrated regression \cite{kuleshov2018accurate}]\label{def:calireg}
  A regression model is well-calibrated if, for any $1 - \alpha \in (0,1)$,
  \begin{equation}
    \Pr\!\left(F_x^{-1}(\alpha/2) \le y \le F_x^{-1}(1-\alpha/2) \,\middle|\, x\right) = 1-\alpha
  \end{equation}
  for every $x$, where $F_x^{-1}$ denotes the inverse CDF of predictive distribution of $y$ given $x$.
\end{definition}

\subsection{Bayesian Learning}\label{sec:bdl}
Bayesian learning is known as a natural and comprehensive framework for UQ, as it incorporates the representation of both aleatoric (data) and epistemic (model) uncertainty. Unlike point estimation methods that rely on a single optimum, i.e., maximum a posteriori (MAP) or maximum likelihood estimation (MLE), Bayesian approaches integrate over all plausible parameter configurations, yielding calibrated predictive uncertainty and improved robustness.

Mathematically, given labeled training dataset $\mathcal{D} = \{(x_i, y_i)\}_{i=1}^N$, and the probabilistic model $p(y \mid x, \theta)$ parameterized by $\theta \in \Theta$, the Bayesian predictive distribution is
\begin{equation}\label{eq:marg}
  p(y \mid x, \mathcal{D}) = \int_\Theta p(y \mid x, \theta)\, p(\theta \mid \mathcal{D})\, \mathrm{d} \theta.
\end{equation}
In practice, this integral is empirically estimated by sampling $\theta$ from the posterior distribution $p(\theta \mid \mathcal{D})$.
The point predictions and the confidence intervals of each prediction are acquired by computing the quantiles of the Bayesian predictive distribution $p(y \mid x,\mathcal{D})$.
Neural networks that predict under this paradigm are called Bayesian neural networks (BNNs) \cite{mackay1992practical, neal1996bayesian,kendall2017uncertainties}.

\begin{definition}[Point Predictions and Credible Intervals for Regression]\label{def:pred}
1) The \emph{point prediction} is the expected value of $y$ under the predictive distribution,
\begin{equation}\label{eq:point_pred}
  \hat{y}(x) = \mathbb{E}_{p(y \mid x,\mathcal{D})}[y]
  = \int y \, p(y \mid x, \mathcal{D}) \, \mathrm{d}y.
\end{equation}
2) Let $F_x^{-1}$ denote the inverse CDF of $p(y\mid x,\mathcal{D})$, which is available in numerical form, a \emph{$(1-\alpha)$ credible interval} for $y$ is obtained by taking the lower $(\alpha/2)$ and upper $(1-\alpha/2)$ quantiles:
\begin{equation}\label{eq:credible_interval}
\mathcal{C}_{1-\alpha}(x) =
\bigl[\,F_x^{-1}(\tfrac{\alpha}{2}),\; F_x^{-1}(1-\tfrac{\alpha}{2})\,\bigr],
\end{equation}
\end{definition}

Under this Bayesian framework, it is also natural to consider the decomposition of the aleatoric and epistemic uncertainty to guarantee the mathematical rigor of UQ \cite{hullermeier2021aleatoric}. Specifically, the aleatoric uncertainty is depicted by the probabilistic model (likelihood) $p(y\mid x,\theta)$ itself, while the epistemic uncertainty is from the posterior $p(\theta \mid \mathcal{D})$.

Posterior inference lies at the core of Bayesian learning.
Specifically, given a dataset $\mathcal{D} = \{(x_i, y_i)\}_{i=1}^N$, a likelihood function $p(y \mid x, \theta)$, and a prior distribution $p(\theta)$, the objective is to infer the posterior distribution over model parameters:
\[
p(\theta \mid \mathcal{D}) = \frac{p(\mathcal{D} \mid \theta)\, p(\theta)}{p(\mathcal{D})},
\]
where $p(\mathcal{D}) = \int p(\mathcal{D} \mid \theta)\, p(\theta)\, d\theta$ is the evidence.
By maintaining a full posterior distribution rather than a single point estimate, Bayesian learning naturally captures epistemic uncertainty,
which can then be propagated to the predictive distribution $p(y \mid x)$.
Such uncertainty quantification serves as the foundation for the label-efficient strategies developed in this work.

In practice, posterior inference implements this process computationally by approximating the intractable posterior distribution, as the marginal likelihood $p(\mathcal{D})$ cannot be computed analytically. Typically, the posterior $p(\theta \mid \mathcal{D})$ is approximated by minimizing the KL divergence between a proposed distribution $q(\theta)$ and the true posterior distribution
\begin{equation}\label{eq:bi_obj}
\underset{q}{\arg\min} \, \mathbb{D}_\mathrm{KL}(q(\theta)\|p(\theta \mid \mathcal{D})).
\end{equation}
The proposed distribution $q(\theta)$ denotes the variational approximation in variational inference, and corresponds to the law of particles in sampling-based approaches.

\subsection{Connections Between Sampling and Finite-Horizon Stochastic Optimal Control}
Bayesian inference can be performed through sampling, where samples drawn from $p(\theta \mid \mathcal{D})$ are used to approximate expectations under the posterior distribution.
Compared to explicit variational methods, sampling-based approaches make fewer assumptions on the posterior and can achieve higher-fidelity approximations of complex posterior distributions \cite{foong2020expressiveness}.
However, sampling from complex and unnormalized high-dimensional posteriors remains a challenging problem in practice.
Recent developments in optimal transport and stochastic control \cite{chen2016relation} have inspired a unifying view of sampling as a Schrödinger bridge problem (\Cref{def:sbp}), naturally formulated as a finite-horizon optimal control problem.

\begin{definition}[Schrödinger bridge problem \cite{leonard2013survey}]\label{def:sbp}
Let $\mathbb{S}$ be a reference path measure on $C([0,1], \mathbb{R}^d)$ (e.g., Wiener measure), and let $\pi_0$, $\pi_1$ be probability measures on $\mathbb{R}^d$. For any path measure $\mathbb{Q}$ on this space, denote by $\mathbb{Q}_t$ its marginal distribution at time $t$. Define the set of admissible path measures:
\[
\mathcal{D}(\pi_0, \pi_1) := \left\{ \mathbb{Q} \ll \mathbb{S} \;\middle|\; \mathbb{Q}_0 = \pi_0,\; \mathbb{Q}_1 = \pi_1 \right\}.
\]
Then the Schrödinger bridge problem seeks
\begin{equation}\label{eq:sbp}
\mathbb{Q}^* = \underset{\mathbb{Q} \in \mathcal{D}(\pi_0, \pi_1)}{\arg\min} \mathbb{D}_{\mathrm{KL}}(\mathbb{Q} \,\|\, \mathbb{S}),
\end{equation}
where $\mathbb{D}_{\mathrm{KL}}$ denotes the Kullback–Leibler (KL) divergence (or relative entropy) on path space.
\end{definition}

\Cref{def:sbp} states the modern formulation of the Schrödinger bridge problem, which seeks the stochastic evolution $\mathbb{Q}$ that transports $\pi_0$ to $\pi_1$ while minimizing the relative entropy with respect to the reference diffusion $\mathbb{S}$. Next, we explain how the posterior inference problem can be reformulated as such a Schrödinger bridge problem, where the key step relies on the data processing inequality (Lemma~\ref{lemma:dpi}).

\begin{lemma}[Data processing inequality (See Appendix~A in
\cite{leonard2013survey})]\label{lemma:dpi}
Let $\mathbb{Q}, \mathbb{S}$ be two probability measures on path space, with terminal distributions $\mu_1 = \mathbb{Q}_1$, $\pi_1 = \mathbb{S}_1$. Then
\[
\mathbb{D}_\mathrm{KL}(\mu_1 \,\|\, \pi_1)
\;\le\;
\mathbb{D}_\mathrm{KL}(\mathbb{Q} \,\|\, \mathbb{S}),
\]
where the equality holds if and only if
\[
\mathbb{Q}(\cdot \mid \theta_1) = \mathbb{S}(\cdot \mid \theta_1) \quad \text{for } \mu_1\text{-almost every } \theta_1.
\]
\end{lemma}

By the data processing inequality, the KL divergence between marginals $\mathbb{D}_\mathrm{KL}(\mu_1 \,\|\, \pi_1)$ is upper-bounded by the path-space KL divergence $\mathbb{D}_\mathrm{KL}(\mathbb{Q} \,\|\, \mathbb{S})$.
Revisit the typical optimization problem \ref{eq:bi_obj} of posterior inference, and let $\mu_1=q(\theta)$ and $\pi_1=p(\theta \mid \mathcal{D})$. The objective of \ref{eq:bi_obj} can be relaxed into a path-space KL minimization problem as formulated in (\ref{eq:sbp}).
This relaxation enables a more tractable and smooth optimization via drift control, as elaborated in the following part.

\begin{proposition}[Stochastic optimal control formulation of Schrödinger bridge problem\cite{leonard2013survey}]\label{theo:socsbp}
Let the reference process $\mathbb{S} = \mathbb{Q}^{f, \pi_0}$ be an Itô diffusion with drift $f$ and constant diffusion coefficient $\gamma$:
\[
\mathrm{d} \theta_t = f(t, \theta_t)\, \mathrm{d}t + \sqrt{\gamma}\,\mathrm{d}B_t,
\qquad \theta_0 \sim \pi_0,
\]
and the controlled process $\mathbb{Q}^{f+u, \pi_0}$
with additional drift control $u$:
\[
\mathrm{d}\theta_t = f(t,\theta_t)\,\mathrm{d}t + u(t,\theta_t)\,\mathrm{d}t + \sqrt{\gamma}\,\mathrm{d}B_t,
\qquad \theta_0 \sim \pi_0.
\]
The Schrödinger bridge problem with marginals $\theta_0 \sim \pi_0$ and $\theta_1 \sim \pi_1$
is equivalent to the finite-horizon stochastic optimal control problem
\begin{equation}
\begin{aligned}
  \min_{u}
  \; &\mathbb{E}\!\left[\int_0^1 \frac{1}{2\gamma}\,\|u(t,\theta_t)\|^2 \,\mathrm{d}t\right],
  \quad \\
  \text{s.t.}\quad
  &\mathrm{d}\theta_t = f(t,\theta_t)\,\mathrm{d}t + u(t,\theta_t)\,\mathrm{d}t + \sqrt{\gamma}\,\mathrm{d}B_t,\;\\
  &\theta_0 \sim \pi_0,\; \theta_1 \sim \pi_1.
\end{aligned}
\label{eq:soc}
\end{equation}
\end{proposition}

\begin{figure*}[ht]
\centering
\includegraphics[width=1\linewidth]{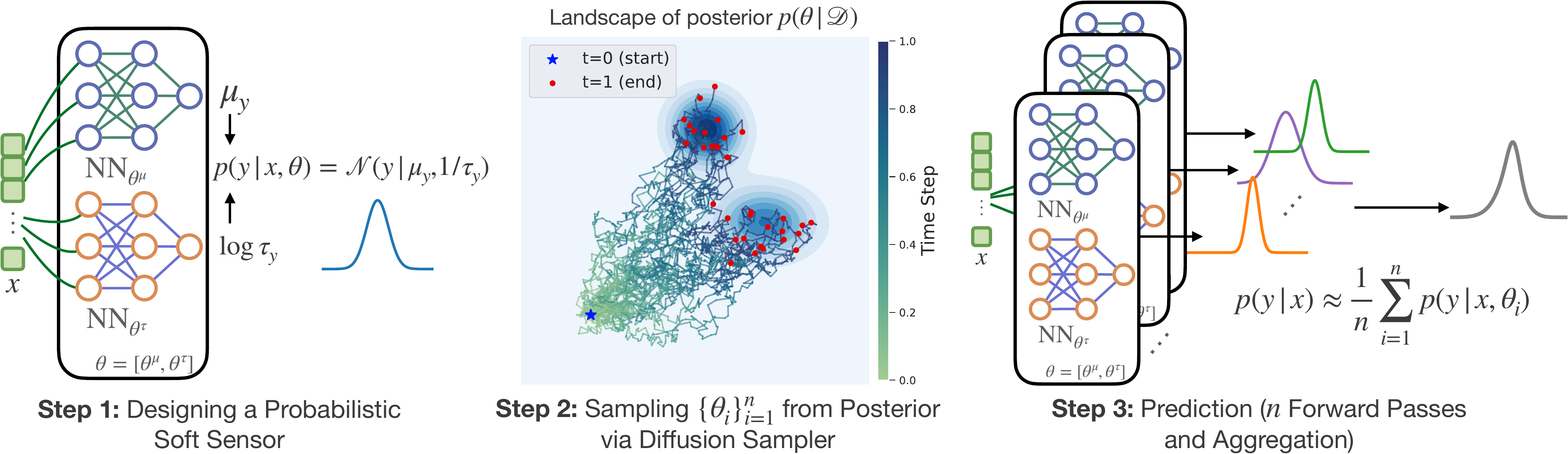}
\caption{Overview of the DiffUQ Framework}
\label{fig:main}
\end{figure*}

The solution of (\ref{eq:soc}) defines a controlled SDE whose simulation yields terminal states distributed according to the target posterior.
This observation motivates the construction of \emph{diffusion samplers}, which approximate the optimal control and generate samples by simulating the controlled SDE and collecting the terminal states.

However, (\ref{eq:soc}) enforces exact matching of the target distribution within a finite time horizon, which can require large control magnitudes when the target is complex, as is often the case for posteriors in industrial models, leading to an ill-conditioned optimization problem. Therefore, we relax this hard constraint into a soft terminal penalty, yielding the following equivalent formulation.

\begin{proposition}[Stochastic optimal control cost with soft terminal penalty \cite{tzen2019theoretical, vargas2023bayesian, zhang2021path}]\label{prop:obj}
Under the conditions $f(t,\theta_t) \equiv 0$ and the initial distribution being the Dirac measure at the origin $\pi_0=\delta_0$, (\ref{eq:soc}) admits, by Girsanov’s theorem and Itô’s formula, the same solution as the following optimization problem:
\begin{equation}
\begin{aligned}
  u^*=&\underset{u}{\arg\min}\;\mathbb{E}\left[ \int_0^1\frac{1}{2\gamma} \|u(t,\theta_t)\|^2 \mathrm{d}t-\log \frac{\pi_1(\theta_1)}{\mathcal{N}(\theta_1|0,\gamma \mathbf{I}_d)}\right], \\
  \text{s.t.}\quad
  &\mathrm{d}\theta_t = u(t,\theta_t)\,\mathrm{d}t + \sqrt{\gamma}\,\mathrm{d}B_t,\;\\
  &\theta_0 \sim \delta_0.
\end{aligned}
\label{eq:obj}
\end{equation}
\end{proposition}

Proposition \ref{prop:obj} converts the hard terminal constraint to a soft terminal penalty, which facilitates a principled trade-off between staying close to the reference prior (running cost) and matching the target posterior distribution (terminal cost).
Moreover, this transition yields a smoother, unconstrained differentiable objective \cite{tzen2019theoretical} that enables end-to-end training via gradient descent.
In practice, the drift control $u(t,\theta_t)$ is parameterized by a neural network and optimized via gradient descent on (\ref{eq:obj}).
Diffusion-based samplers, such as the Path Integral Sampler (PIS) \cite{zhang2021path} and the Neural Schrödinger–Föllmer Sampler (NSFS) \cite{vargas2023bayesian}, build upon this relaxed formulation. Both methods learn to approximate an optimal control policy with neural networks, but differ in their theoretical derivations, parameterization schemes, and application domains.

\section{Uncertainty Quantification in Industrial Models via Diffusion Sampler}\label{sec:method}
This section elaborates on how to incorporate diffusion samplers into industrial data-driven modeling to equip models with reliable and well-calibrated predictive uncertainty. Given that the diffusion sampler lies at the core of modeling epistemic uncertainty in our framework, we refer to our method as Diffusion-based Uncertainty Quantification (DiffUQ).

\Cref{fig:main} illustrates the overall workflow of our proposed method. Specifically, our method includes 3 steps: 1) construct a probabilistic regression model $p(y \mid x, \theta)$ as the base model, where $x$ denotes the process variables, $y$ is the target variable to be predicted, and $\theta$ represents the model's parameter vector; 2) draw $n$ samples $\{\theta_i\}_{i=1}^n$ from the posterior $p(\theta \mid \mathcal{D})$ given a dataset $\mathcal{D}$; 3) predict by $n$ forward passes and aggregation.

Epistemic uncertainty arises from the presence of multiple plausible parameter configurations, which are captured by the complex and typically unnormalized posterior distribution $p(\theta \mid \mathcal{D})$ after observing the dataset $\mathcal{D}$.

\subsection{Probabilistic Regression Model}\label{sec:model}
The probabilistic model captures the conditional distribution of $y$ given $x$ and $\theta$, thereby providing a framework to predict the target variable and represent aleatoric uncertainty.
A proper probabilistic model $p(y\mid x,\theta)$ is critical for both accurate prediction and aleatoric uncertainty representation. Considering that industrial processes may have heteroscedastic data noise \cite{kay2022integrating}, industrial models require explicit estimation of input-dependent noise variances.
Consequently, two distinct neural networks (or, more generally, parametric functions) are employed to parameterize the probabilistic model, providing the mean and precision, respectively.
\begin{equation}\label{eq:model}
p(y|x,\theta) = \mathcal{N}(\mathrm{NN}_{\theta^\mu}(x), e^{- \mathrm{NN}_{\theta^\tau}(x)}),
\end{equation}
where $\theta = [\theta^\mu, \theta^\tau]$. Specifically, $\theta^{\mu}$ and $\theta^{\tau}$ parameterize the mean network $\mathrm{NN}_{\theta^{\mu}}$ (point prediction) and the precision network $\mathrm{NN}_{\theta^{\tau}}$ (aleatoric uncertainty), respectively. This design enables learning input-dependent noise variance instead of using a fixed one. Note that despite the notation `'$\mathrm{NN}$', these components can be any differentiable parametric functions.

\subsection{Posterior Sampling via Diffusion Sampler}
\begin{algorithm}[tp]
\caption{Training a diffusion sampler}
\label{alg:ds}
\begin{algorithmic}[1]
\REQUIRE Dataset $\mathcal{D}=\{x_i, y_i\}_{y=1}^N$, probabilistic model $p(\cdot|x,\theta)$, prior $p(\theta)= \mathcal{N}(0, I)$.
% $\left(p(\mathcal{D}|\theta) = \prod_{i=1}^N p(y_i|x_i,\theta)\right)$
\ENSURE $u_{\phi}(t,\theta)$ parameterized by $\phi$

\STATE \textbf{Define:} Augmented SDE drift
$f_\phi(t, [\theta_t, c_t])
= \big[ u_\phi(t,\mathbf{\theta}_t), \tfrac{1}{2}\|u_\phi(t,\mathbf{\theta}_t)\|^2 \big]$,
diffusion $g(t, [\theta_t, c_t]) = [\gamma,0]$, $\theta_0 = \textbf{0}$, $c_0 = 0$; discretization step $\Delta t = 1/T$

\FOR{i = 1 \textbf{to} max\_iter}
\STATE SDE simulation
\begin{equation}
\begin{aligned}
(\theta_1, c_1)
\leftarrow & (\theta_0, c_0)
+ \Sigma_{t=0}^T f\!\big(t, [\theta_t, c_t]\big)\, \Delta t \\
&+ \Sigma_{t=0}^T g\!\big(t, [\theta_t, c_t]\big)\, \Delta W_t
\end{aligned}\label{eq:em}
\end{equation}
\STATE Sample mini-batch $\mathcal{S} \subset \mathcal{D}$ and run gradient descent step
\begin{align*}
\phi \leftarrow \phi - \eta \;\nabla_\phi \big[c_1 - \tfrac{|\mathcal{D}|}{|\mathcal{S}|}\log p(\mathcal{S}|\theta_1)p(\theta_1) \\
+ \log{\mathcal{N}(\theta_1|0,\gamma \mathbf{I}_d)} \big]
\end{align*}
\ENDFOR
\end{algorithmic}
\end{algorithm}
\subsubsection{Training} With the probabilistic regression model $p(y \mid x, \theta)$ defined in \cref{sec:model} and a standard Gaussian distribution as an uninformative prior, we utilize a diffusion sampler to obtain samples from the resulting parameter posterior given $\mathcal{D}$. \Cref{alg:ds} shows the complete training procedure of the training process. Consider a reference Wiener process $\mathrm{d}\theta = \gamma\,\mathrm{d}W_t$. Instead of directly working with $\theta$, the goal is to learn a optimal control $u_\phi(t,\theta)$ parameterized by $\phi$, such that the terminal states induced by the controlled diffusion process $\mathrm{d}\theta = u_\phi(t,\theta)\,\mathrm{d}t + \gamma\,\mathrm{d}W_t$ distribute according to the posterior $p(\theta|\mathcal{D})$. This goal can be accomplished through gradient descent on a single objective (\ref{eq:obj}). An augmented state $c_t$ is used to integrate the running cost term $\|u\|^2$ along the path:
$$
c_1 = \int_0^1 \frac{1}{2\gamma} \| u_\phi(t, \theta_t) \|^2 \mathrm{d}t
$$
thus significantly reducing peak memory usage during training.

The drift neural network $u_\phi(t,\theta)$ is parameterized\footnote{
Standard SDE theory assumes Lipschitz continuity and linear growth of the drift, which are not explicitly enforced for drift parameterizations. In practice, GELU and layer normalization can promote stable dynamics. The diffusion mechanism further encourages smooth optimization landscapes (see \Cref{fig:training}).
} as an 8-layer MLP of width 32, using GELU activations for all experiments in our work (see \Cref{apdx:capacity} for details on network capacity). Layer normalization without learnable affine parameters is also adopted, as it can stabilize training. Batch normalization is strictly forbidden as it introduces dependence between samples within a batch and compromises the uniqueness of the solution \cite{xu2022infinitely}.

\Cref{eq:em} shows that the SDE is simulated using the Euler–Maruyama (EM) discretization scheme, where $\Delta W_t \sim \mathcal{N}(0,\Delta t)$. Theoretically, this discretization has a provable error bound \cite{huang2021schrodinger, vargas2023bayesian} for finite-horizon stochastic optimal control problems. This discrete SDE simulation is implemented in an efficient manner using \texttt{torchsde} \cite{li2020scalable, kidger2021neuralsde}, which integrates $n$ trajectories in parallel. We use $\Delta t=0.04$ for the training process and $\Delta t = 0.01$, which will be introduced next.

\subsubsection{Sampling} After training, $n$ i.i.d.\ samples $\{\theta_i\}_{i=1}^n$ are generated via simulating $n$ trajectories and collecting the their terminal states. Each sample requires only $1/\Delta t$ forward passes of the control network, and all trajectories are computed in parallel via \texttt{torchsde}. Hence, the per-sample computational cost becomes marginal and can be regarded as amortized into the training phase.

In our framework, multiple forward predictions are indeed used to estimate predictive statistics. However, the key distinction is that the diversity of posterior samples is not primarily governed by the number of samples drawn at inference time, but by the learned diffusion dynamics. During training, the controlled SDE learns a global transport that maps the reference distribution to the posterior manifold, effectively encoding multi-modal structure in the drift function. As a result, even a small number of trajectories can traverse different modes, since exploration is embedded in the dynamics itself rather than achieved through independent Monte Carlo exploration. Increasing the number of samples mainly reduces Monte Carlo estimation variance, but does not fundamentally alter posterior coverage.

\subsection{Posterior Prediction}
Once posterior samples are obtained, the posterior predictive distribution $p(y \mid x)$ is evaluated via the posterior predictive equation (\ref{eq:marg}), where the intractable integral is approximated empirically using Monte Carlo estimation over the posterior samples
\begin{equation}\label{eq:emprical}
p(y\mid x,\mathcal{D})=\frac{1}{n} \sum_{i=1}^n p(y\mid x, \theta_i),
\end{equation}
where $\theta_i \sim p(\theta \mid \mathcal{D}), i=1,...,n.$
\Cref{eq:emprical} implies that each prediction of our method requires $n$ forward passes, which, in our implementation, is parallelized using \texttt{torch.vmap}. This process is referred to as aggregation and is also known as an ensemble of base models in some literature. Finally, in accordance with \cref{def:pred}, uncertainty-aware prediction is performed.

\section{Experiments}
In this section, we first introduce the evaluation setup (\Cref{sec:eval}), including the metrics and baseline methods. We then demonstrate the advantage of diffusion-based sampling in capturing complex posterior distributions of industrial models through two illustrative toy examples (\Cref{sec:toy}). Next, the effectiveness of DiffUQ is validated on a simple linear Raman-based soft sensor from the penicillin fermentation process (\Cref{sec:pensim}). Finally, experiments on a neural network–based industrial process model for the real-world ammonia synthesis process further highlight the capability of DiffUQ to deliver reliable uncertainty quantification in higher-dimensional parameter spaces (\Cref{sec:hlt}).
\subsection{Evaluation Setup}\label{sec:eval}
\subsubsection{Metrics}
Consider a test set $\mathcal{D}_\mathrm{test}=\{(x_i,y_i)\}_{i=1}^M$ of size $M$. A series of evaluation metrics is employed to assess both the point prediction accuracy and the predictive uncertainty quality. To jointly evaluate these two aspects, the mean negative log-likelihood (NLL) is adopted to measure the quality of probabilistic predictions:
\begin{equation}
\mathrm{NLL} = -\frac{1}{M} \sum_{i=1}^{M} \log p(y_i \mid x_i, \mathcal{D}),
\end{equation}
where $p(y_i \mid x_i, \mathcal{D})$ denotes the posterior predictive distribution given by the uncertainty-aware model. A lower NLL indicates that the model produces confident and accurate predictions by jointly accounting for accuracy and uncertainty calibration.

To further evaluate the calibration of predictive uncertainty, the expected calibration error (ECE) and maximum calibration error (MCE) \cite{kuleshov2018accurate, naeini2015obtaining} are computed based on the empirical coverage of credible intervals. Given the posterior predictive distribution (\ref{eq:emprical}), the empirical coverage level $\hat{p}$ of the nominal coverage level $p$ is defined as
$
\hat{p} = \frac{1}{M} \sum_{j=1}^{M} \mathbb{I}\{y_i \in \mathcal{C}_{p} (x_i)\},
$
where $\mathcal{C}_{p} (x_i)$ is the credible interval defined in (\ref{eq:credible_interval}).
The expected calibration error is then given by
\begin{equation}
\mathrm{ECE} = \frac{1}{B} \sum_{k=0}^{B} |\hat{p}_k - p_k|,
\end{equation}
while the maximum calibration error is defined as
\begin{equation}
\mathrm{MCE} = \max_j(|\hat{p}_k - p_k|),
\end{equation}
where $p_k = k/B$, $k=1,\ldots,B$, and $B$ denotes the number of different nominal coverage levels evaluated. ECE evaluates the average alignment between predicted confidence and empirical accuracy, reflecting overall reliability. MCE captures the worst-case calibration error, which is crucial for preventing dangerous overconfidence in safety-critical operations. Together, they validate the trustworthiness of the predictive distribution.

For point prediction accuracy, denoting $\hat{y}_i$ the point prediction (\ref{eq:point_pred})  given $x_i$, the mean squared error (MSE) and mean absolute error (MAE) are used:
\begin{equation}
\text{MSE} = \frac{1}{M} \sum_{i=1}^{M} (y_i - \hat{y}_i)^2,
\quad
\text{MAE} = \frac{1}{M} \sum_{i=1}^{M} |y_i - \hat{y}_i|.
\end{equation}
Finally, the coefficient of determination ($R^2$) evaluates the overall goodness of fit:
\begin{equation}
R^2 = 1 - \frac{\sum_{i=1}^{M} (y_i - \hat{y}_i)^2}{\sum_{i=1}^{M} (y_i - \bar{y})^2},
\quad \text{where } \bar{y} = \frac{1}{M} \sum_i y_i.
\end{equation}

\subsubsection{Baseline Methods}
We compare our approach with several representative uncertainty quantification methods that cover the main methodological categories commonly adopted in industrial data-driven modeling. Specifically, we include:
(i) MC Dropout \cite{yang2022remaining, cao2023parallel}, which approximates Bayesian inference by injecting stochasticity at test time;
(ii) deep ensembles (DE) \cite{lakshminarayanan2017simple}, which capture model uncertainty through diversity across independently trained networks;
(iii) mean-field variational inference (MFVI) \cite{lee2021uncertainty}, a classical variational Bayesian approach with factorized posterior assumptions;
(iv) stochastic gradient Langevin dynamics (SGLD), as a representative stochastic-gradient MCMC method \cite{welling2011bayesian}; and
(v) Stein variational gradient descent (SVGD) \cite{ma2024heat}, representing deterministic particle-based Bayesian inference.

These methods are selected because they represent widely used and conceptually distinct approaches to uncertainty quantification, spanning variational inference, ensemble-based, and particle-based variational inference. In addition, we report results obtained using maximum a posteriori (MAP) estimation. Although MAP yields only a point estimate of model parameters and therefore does not account for model uncertainty, data uncertainty is still captured through the probabilistic output model, making MAP a meaningful baseline for assessing the effect of explicitly modeling parameter uncertainty.

In order to enable a comparison of the intrinsic calibration accuracy of uncertainty estimates, all methods are evaluated without post-hoc calibration.

\subsection{Sampling from Toy Posterior Distributions}\label{sec:toy}
This section presents synthetic examples that demonstrate the diffusion sampler's capacity to explore complex or even ill-posed posterior distributions. In particular, the non-Gaussian and multimodal `smiley-face' distribution and an ill-posed funnel distribution featuring a narrow neck and wide base are examined, reflecting typical challenges that may arise in posteriors of industrial models.

\begin{figure}[h!]
\centering
\begin{subfigure}[t]{0.45\linewidth}
\centering
\begin{subfigure}[t]{\linewidth}
\centering
\includegraphics[width=0.75\linewidth]{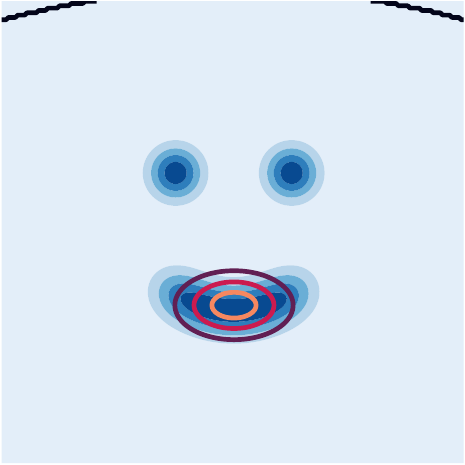}
\caption{MFVI}
\label{fig:smiley_mfvi}
\end{subfigure}
\begin{subfigure}[t]{\linewidth}
\centering
\includegraphics[width=0.75\linewidth]{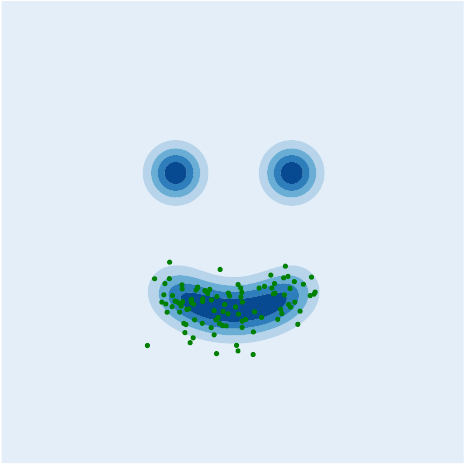}
\caption{SGLD}
\label{fig:smiley_sgld}
\end{subfigure}
\begin{subfigure}[t]{\linewidth}
\centering
\includegraphics[width=0.75\linewidth]{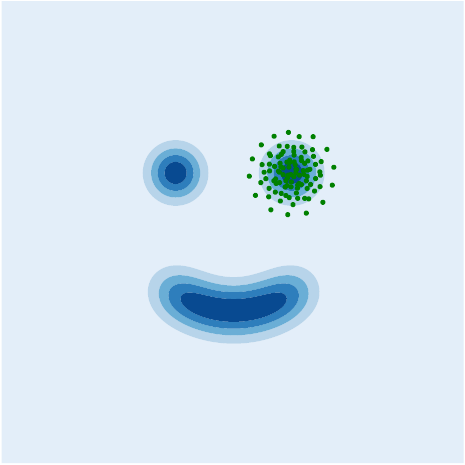}
\caption{SVGD}
\label{fig:smiley_svgd}
\end{subfigure}
\begin{subfigure}[t]{\linewidth}
\centering
\includegraphics[width=0.75\linewidth]{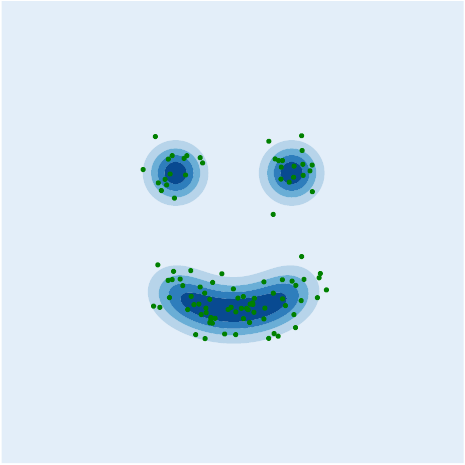}
\caption{\textbf{Diffusion Sampler}}
\label{fig:smiley_diff}
\end{subfigure}
\end{subfigure}\hspace{20pt}
\begin{subfigure}[t]{0.45\linewidth}
\centering
\begin{subfigure}[t]{\linewidth}
\centering
\includegraphics[width=0.75\linewidth]{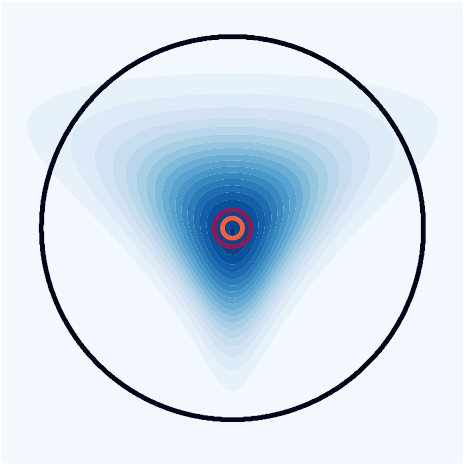}
\caption{MFVI}
\label{fig:funnel_mfvi}
\end{subfigure}
\begin{subfigure}[t]{\linewidth}
\centering
\includegraphics[width=0.75\linewidth]{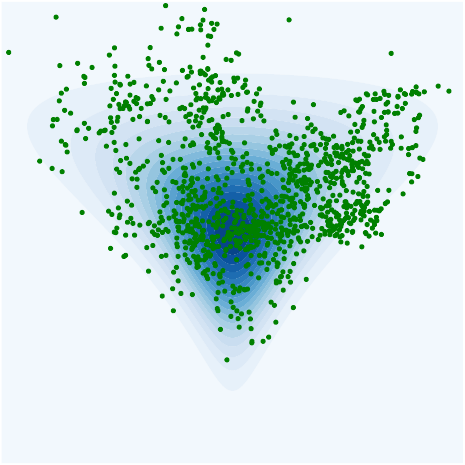}
\caption{SGLD}
\label{fig:funnel_sgld}
\end{subfigure}
\begin{subfigure}[t]{\linewidth}
\centering
\includegraphics[width=0.75\linewidth]{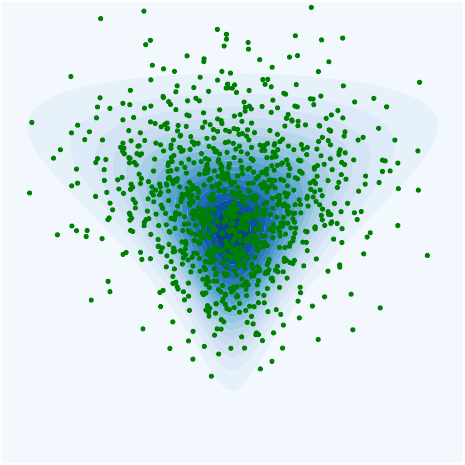}
\caption{SVGD}
\label{fig:funnel_svgd}
\end{subfigure}
\begin{subfigure}[t]{\linewidth}
\centering
\includegraphics[width=0.75\linewidth]{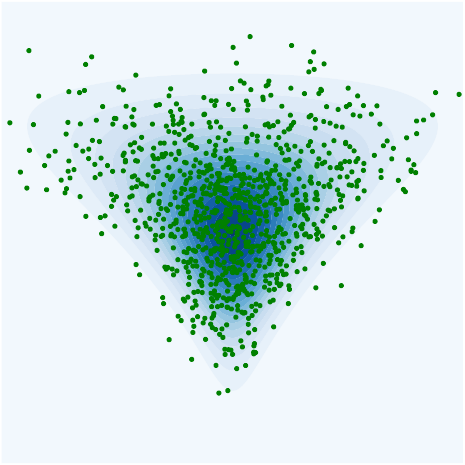}
\caption{\textbf{Diffusion Sampler}}
\label{fig:funnel_diff}
\end{subfigure}
\end{subfigure}
\caption{Comparison on the smiley-face (left) and funnel (right) distributions. (a,e) show mean-field variational contours; other panels show samples from different sampling methods.}
\label{fig:smiley_funnel}
\end{figure}

MFVI restricts the variational approximation to a mean-field Gaussian family, and \Cref{fig:smiley_mfvi}, \ref{fig:funnel_mfvi} illustrate the limitations of this assumption when representing complex target distributions.
Gradient flow–based sampling methods, such as SGLD (\Cref{fig:smiley_sgld}, \ref{fig:funnel_sgld}) and SVGD (\Cref{fig:smiley_svgd}, \ref{fig:funnel_svgd}), both fail to fully recover the structural characteristics of the target distribution.
\Cref{fig:smiley_diff}, \ref{fig:funnel_diff} demonstrates that the diffusion sampler can generate high-quality samples covering all modes of such challenging distributions, outperforming the compared methods.

\subsection{Penicillin Fermentation System Simulation}\label{sec:pensim}

\begin{figure}[ht]
\centering
\includegraphics[width=0.95\linewidth]{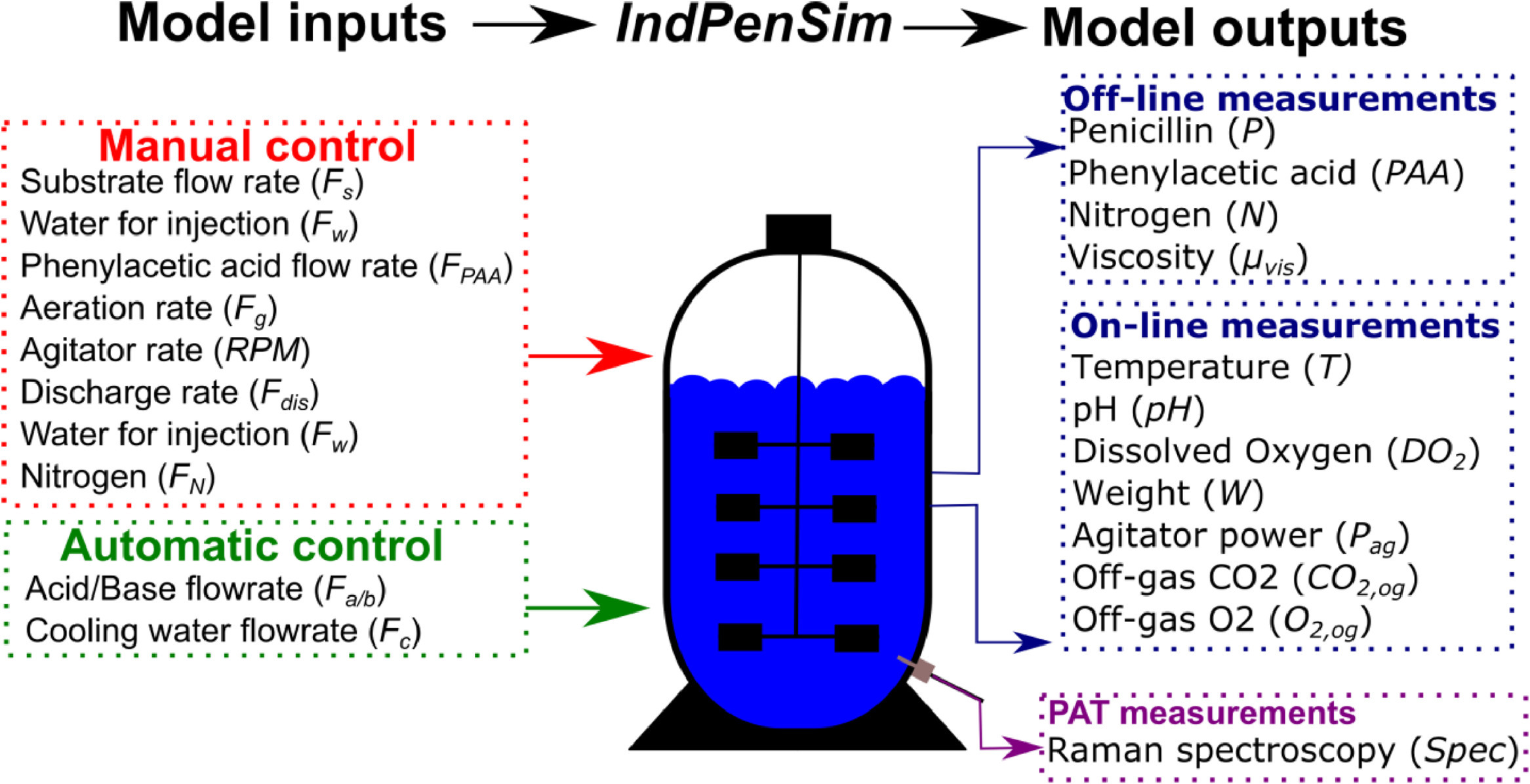}
\caption{Summary of the first principle-based mathematical simulator of industrial-scale penicillin simulation \cite{goldrick2019modern}}
\label{fig:indpensim}
\end{figure}

In this section, we consider an industrial-scale simulation (IndPenSim) \cite{goldrick2019modern}, shown on \Cref{fig:indpensim}, a 100,000-litre penicillin fermentation process. The dataset includes 100 batches of process and Raman spectroscopy measurements, which is the largest available resource for advanced data analytics in this domain. The process involves a Raman-based phenylacetic acid (PAA) soft sensor task, where Raman spectra serve as input variables to predict the PAA concentration.
Since the Raman-based PAA soft sensor exhibits a strong linearity, it serves as an appropriate benchmark to investigate the performance of our UQ method in the context of simple linear models. Hence, this setting provides a simple case that is well-suited for comparing the effectiveness of different methods.

\subsubsection{Task Description} In this case study, the purpose is to develop a Raman-based soft sensor for the online prediction of PAA concentration during penicillin fermentation \cite{goldrick2019modern}. Out of the 100 batches, we select batches 1–60 for training and batches 61–90 for testing. The first 60 batches correspond to open-loop or manually operated processes, while batches 61–90 are subject to closed-loop control based on a simple linear soft sensor of PAA, which reflects the intended application of PAA prediction—supporting closed-loop process control. This split, therefore, ensures that the evaluation directly aligns with the original motivation of the task.

Raman spectra are generated in IndPenSim every 12 minutes across the wavenumber range 250–2250 cm$^{-1}$, yielding a large high-dimensional dataset, which is preprocessed following \cite{goldrick2019modern}. Specifically, the spectral regions 1540–1580 cm$^{-1}$ and 1950–2050 cm$^{-1}$ were identified and selected as informative for PAA concentration.
The selected spectral regions were preprocessed using a Savitzky–Golay smoothing filter (15-point window), followed by taking the first derivative. These processed spectra served as the secondary variables $x$, while offline PAA measurements were interpolated via cubic splines to align with the spectral acquisition times.

\begin{table*}[ht]
\centering
\caption{Results of Raman-based PAA soft sensor.}
\label{tab:pensim}
\begin{threeparttable}
\begin{tabular}{lllllll}
\toprule
&  \multicolumn{3}{l}{Uncertainty Quality} & \multicolumn{3}{l}{Accuracy} \\
\cmidrule(l){2-4} \cmidrule(l){5-7}
Method & NLL & ECE & MCE & MSE & MAE & $R^2$  \\
\midrule
MAP & -2.6339 ± 0.2131 & 0.1331 ± 0.0896 & 0.2267 ± 0.1564 & 524.7643 ± 86.7733 & 18.1418 ± 1.4578 & 0.9020 ± 0.0162 \\
\midrule
DE & -2.8330 ± 0.0085 & 0.0654 ± 0.0044 & 0.1058 ± 0.0068 & 493.4771 ± 6.1154 & 17.6894 ± 0.0907 & 0.9078 ± 0.0011 \\
MC Dropout & -2.3574 ± 0.4713 & 0.1421 ± 0.1027 & 0.2537 ± 0.1963 & 1414.1550 ± 1809.8815 & 26.3602 ± 16.5030 & 0.7358 ± 0.3381 \\
MFVI & -2.8593 ± 0.0238 & 0.0493 ± 0.0071 & 0.0856 ± 0.0111 & 486.7107 ± 26.2902 & 17.5233 ± 0.4535 & 0.9091 ± 0.0049 \\
SGLD & -2.4741 ± 0.0271 & 0.2036 ± 0.0080 & 0.3630 ± 0.0128 & 541.3067 ± 7.5517 & 18.5968 ± 0.1380 & 0.8989 ± 0.0014 \\
SVGD & -2.3324 ± 0.0128 & 0.1111 ± 0.0215 & 0.2046 ± 0.0482 & 1001.8524 ± 83.1500 & 25.6316 ± 1.1111 & 0.8128 ± 0.0155 \\
\textbf{DiffUQ} & \textbf{-2.8741} ± 0.0021 & \textbf{0.0429} ± 0.0009 & \textbf{0.0716} ± 0.0015 & \textbf{477.8259} ± 2.1541 & \textbf{17.3031} ± 0.0379 & \textbf{0.9107} ± 0.0004 \\
\bottomrule
\end{tabular}

\begin{tablenotes}
% \footnotesize
\item $n=64$ posterior samples are used for all Bayesian methods, except MAP. Values are reported as mean $\pm$ standard deviation over five runs. \textbf{Bold} indicates the best result.
\end{tablenotes}
\end{threeparttable}
\end{table*}

Conventional Raman-based PAA soft sensors, however, provide only point predictions without accounting for predictive uncertainty. Considering prediction of PAA concentrations is essential for operator decision-making and closed-loop control, incorporating well-calibrated uncertainty estimates has the potential to enhance downstream task performance and improve overall process robustness and safety.

\subsubsection{Probabilistic Model Structure} The probabilistic regression model for Raman-based PAA soft sensor follows (\ref{eq:model}). Considering the classical Raman-based PAA soft sensor is linear \cite{goldrick2019modern}, $\mathrm{NN}_{\theta^\mu}$ is specified as a linear model. $\mathrm{NN}_{\theta^\tau}$ is specified as an MLP with 1 hidden layer of width 4 to capture data noise.

\subsubsection{Results}
\Cref{tab:pensim} compares the performance of DiffUQ with other uncertainty quantification methods on the Raman-based PAA soft sensor task.
We report both uncertainty quality metrics (NLL, ECE, MCE) and accuracy-oriented metrics ($R^2$, MSE, MAE). DiffUQ consistently outperforms alternative methods: it achieves the lowest NLL, ECE, and MCE, indicating well-calibrated predictive distributions, while simultaneously attaining the best $R^2$, MSE, and MAE, reflecting superior point prediction quality. Although MFVI assumes a factorized Gaussian variational family, this structural bias appears mild in this relatively simple task; in contrast, sampling-based methods may suffer from finite-sample effects and mixing inefficiencies, leading to inferior empirical performance.

The case of the Raman-based PAA soft sensor highlights that our DiffUQ, when applied to a linear model, yields highly competitive performance.

\subsection{Real-World Case Study: High-Low Transformer Unit of Ammonia Synthesis Process}\label{sec:hlt}
\begin{figure}[ht]
\centering
\includegraphics[width=0.95\linewidth]{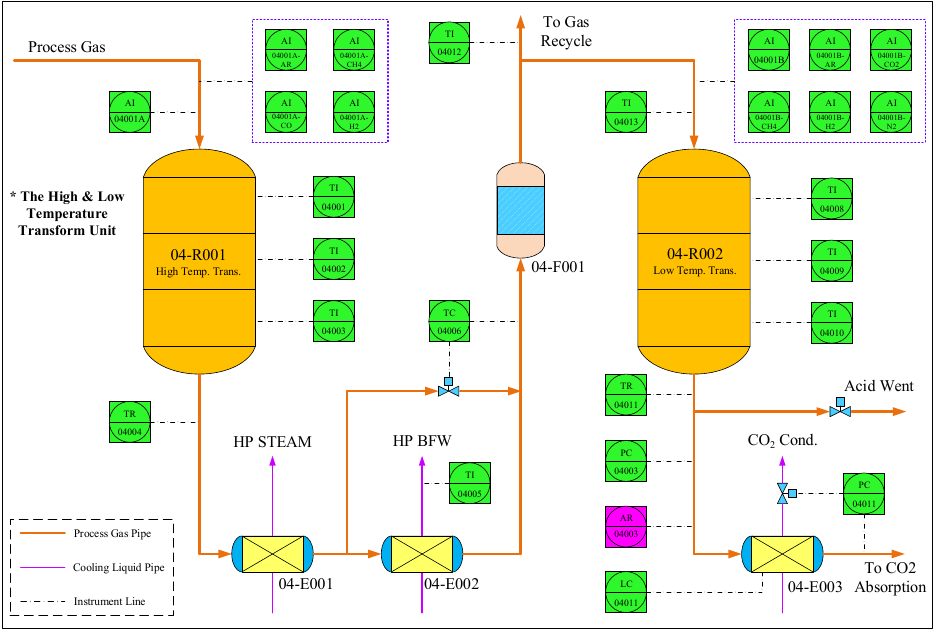}
\caption{High-Low Transformer unit from an ammonia synthesis process.}
\label{fig:hlt}
\end{figure}

This section uses the residual carbon monoxide ($\mathrm{CO}$) concentration prediction task from the high-low transformer (HLT) unit of an ammonia synthesis process as a real-world case study to demonstrate the value of DiffUQ. Unlike using Raman spectra as secondary variables $x$, predicting residual $\mathrm{CO}$ concentrations depends on dynamic process variables and involves nonlinear relationships. Thus, this process model requires a more complex neural network architecture.

\begin{table*}[ht]
\centering
\caption{Uncertainty qualities and accuracy of the HLT residual $\mathrm{CO}$ concentration model based on our method compared with commonly used UQ approaches in industrial models.}\label{tab:hlt}
\begin{threeparttable}
\begin{tabular}{llllllll}
\toprule
&  & \multicolumn{3}{l}{Uncertainty Quality} & \multicolumn{3}{l}{Accuracy} \\
\cmidrule(l){3-5} \cmidrule(l){6-8}
Method & $n$ & NLL & ECE & MCE & MSE & MAE & $R^2$\\
\midrule
MAP & 1 & -0.2729 ± 0.0841 & 0.0321 ± 0.0269 & 0.0649 ± 0.0469 & 0.0330 ± 0.0034 & 0.1440 ± 0.0079 & 0.9430 ± 0.0059 \\
\cmidrule{1-8}
\multirow[t]{3}{*}{DE} & 4 & -0.3363 ± 0.0216 & 0.0210 ± 0.0104 & 0.0404 ± 0.0156 & 0.0298 ± 0.0015 & 0.1365 ± 0.0040 & 0.9472 ± 0.0026 \\
& 64 & -0.3399 ± 0.0204 & 0.0227 ± 0.0101 & 0.0450 ± 0.0210 & 0.0295 ± 0.0015 & 0.1363 ± 0.0037 & 0.9489 ± 0.0036 \\
& 128 & -0.3290 ± 0.0387 & 0.0229 ± 0.0171 & 0.0480 ± 0.0303 & 0.0301 ± 0.0028 & 0.1376 ± 0.0063 & 0.9479 ± 0.0039 \\
\cmidrule{1-8}
\multirow[t]{3}{*}{MC Dropout} & 4 & -0.0038 ± 0.0100 & 0.0758 ± 0.0017 & 0.1364 ± 0.0049 & 0.0602 ± 0.0017 & 0.1920 ± 0.0021 & 0.8959 ± 0.0029 \\
& 64 & -0.0296 ± 0.0019 & 0.1059 ± 0.0010 & 0.1847 ± 0.0038 & 0.0499 ± 0.0002 & 0.1769 ± 0.0005 & 0.9138 ± 0.0004 \\
& 128 & -0.0291 ± 0.0004 & 0.1090 ± 0.0005 & 0.1881 ± 0.0019 & 0.0498 ± 0.0001 & 0.1766 ± 0.0002 & 0.9138 ± 0.0001 \\
\cmidrule{1-8}
\multirow[t]{3}{*}{MFVI} & 4 & -0.0592 ± 0.1246 & 0.0910 ± 0.0201 & 0.1614 ± 0.0398 & 0.0379 ± 0.0044 & 0.1557 ± 0.0103 & 0.9344 ± 0.0075 \\
& 64 & -0.0611 ± 0.1247 & 0.0906 ± 0.0203 & 0.1608 ± 0.0400 & 0.0379 ± 0.0043 & 0.1556 ± 0.0102 & 0.9345 ± 0.0075 \\
& 128 & -0.0614 ± 0.1249 & 0.0905 ± 0.0203 & 0.1609 ± 0.0398 & 0.0379 ± 0.0043 & 0.1556 ± 0.0102 & 0.9345 ± 0.0075 \\
\cmidrule{1-8}
\multirow[t]{3}{*}{SGLD} & 4 & \underline{-0.3666} ± 0.0094 & 0.0186 ± 0.0053 & 0.0376 ± 0.0084 & \underline{0.0280} ± 0.0006 & \underline{0.1325} ± 0.0015 & \underline{0.9515} ± 0.0010 \\
& 64 & \underline{-0.3734} ± 0.0032 & 0.0199 ± 0.0009 & 0.0412 ± 0.0012 & \underline{0.0276} ± 0.0002 & \underline{0.1315} ± 0.0005 & \underline{0.9522} ± 0.0003 \\
& 128 & \underline{-0.3714} ± 0.0038 & 0.0195 ± 0.0011 & 0.0401 ± 0.0007 & \underline{0.0278} ± 0.0002 & \underline{0.1319} ± 0.0006 & \underline{0.9520} ± 0.0004 \\
\cmidrule{1-8}
\multirow[t]{3}{*}{SVGD} & 4 & -0.3516 ± 0.0157 & \textbf{0.0063} ± 0.0035 & \textbf{0.0132} ± 0.0058 & 0.0290 ± 0.0009 & 0.1347 ± 0.0024 & 0.9498 ± 0.0016 \\
& 64 & -0.3697 ± 0.0092 & \underline{0.0134} ± 0.0027 & \underline{0.0272} ± 0.0057 & 0.0279 ± 0.0005 & 0.1321 ± 0.0014 & 0.9517 ± 0.0009 \\
& 128 & -0.3683 ± 0.0114 & \underline{0.0125} ± 0.0040 & \underline{0.0249} ± 0.0068 & 0.0280 ± 0.0007 & 0.1323 ± 0.0017 & 0.9515 ± 0.0011 \\
\cmidrule{1-8}
\multirow[t]{3}{*}{\textbf{DiffUQ}} & 4 & \textbf{-0.4262} ± 0.0073 & \underline{0.0088} ± 0.0028 & \underline{0.0237} ± 0.0034 & \textbf{0.0250} ± 0.0003 & \textbf{0.1234} ± 0.0010 & \textbf{0.9568} ± 0.0006 \\
& 64 & \textbf{-0.4242} ± 0.0064 & \textbf{0.0081} ± 0.0011 & \textbf{0.0233} ± 0.0010 & \textbf{0.0251} ± 0.0003 & \textbf{0.1237} ± 0.0009 & \textbf{0.9566} ± 0.0005 \\
& 128 & \textbf{-0.4266} ± 0.0055 & \textbf{0.0086} ± 0.0022 & \textbf{0.0238} ± 0.0021 & \textbf{0.0250} ± 0.0003 & \textbf{0.1234} ± 0.0007 & \textbf{0.9568} ± 0.0004 \\
\bottomrule
\end{tabular}
\begin{tablenotes}
% \footnotesize
\item Values are reported as `mean $\pm$ standard deviation' over five runs. \textbf{Bold} indicates the best result, and \underline{underline} indicates the second best.
\end{tablenotes}
\end{threeparttable}
\end{table*}

\subsubsection{Task Description}
The ammonia synthesis process serves as a representative industrial benchmark: it is operated at large scale under extreme conditions, involves strongly coupled and nonlinear unit operations, and faces safety-critical and economically vital constraints. The key process variables (e.g., residual $\mathrm{CO}$ concentration of HLT unit) are difficult to measure online, motivating the use of data-driven models.

The dataset employed in this case study originates from the high-low transformer (HLT) unit of a full-scale ammonia synthesis plant, as illustrated in \Cref{fig:hlt}.
$\mathrm{CO}$ is a severe poison for ammonia synthesis catalysts, and excessive $\mathrm{CO}$ also leads to unnecessary $\mathrm{H_2}$ consumption during downstream methanation.
Controlling the $\mathrm{CO}$ concentration at the outlet of the HLT unit is crucial to protect the catalyst and maintain the efficiency of the process.
However, this key quality variable is not directly available via online measurement but is instead obtained offline through gas chromatography, hindering the timely intervention.

The objective is to develop a process model capable of estimating the residual $\mathrm{CO}$ concentration in real time, using the available process measurements, thereby eliminating the reliance on delayed laboratory assay results. Moreover, credible prediction intervals are as operationally important as the predicted residual $\mathrm{CO}$ value itself, as they guide risk-aware operational decision making about timely intervention and enhance process safety.
The collected data contains 26 continuously monitored operational measurements, including flow rates, inlet gas compositions, temperatures, pressures, and liquid levels, as well as one key variable measured offline: the residual $\mathrm{CO}$ concentration at the unit outlet. To account for temporal dependencies within the process, the dataset is split into training and testing subsets based on the sampling order, with the initial 80\% of continuous samples for training and the final 20\% for evaluation.

\subsubsection{Probabilistic Model Structure}
The probabilistic regression model in this case study follows (\ref{eq:model}), where $\mathrm{NN}_{\theta_1}$ is a 3-layer MLP with a hidden layer of 32 neurons, and $\mathrm{NN}_{\theta_2}$ is a 4-layer MLP with 2 hidden layers of width 4 and 2.

\subsubsection{Main Results} \Cref{tab:hlt} reports the uncertainty quality and accuracy of our method against the reference methods. Except for MAP, which only finds a single optimal parameter vector, all approximate inference methods are evaluated with different sampling sizes of 4, 64, and 128.

In comparison with commonly used approximate inference methods, DiffUQ achieves consistently superior uncertainty metrics across different sample sizes. An exception is observed for SVGD at $n=4$, where calibration is slightly better.
However, this is accompanied by reduced predictive accuracy ($R^2$/MSE/MAE) and a worse NLL. These results suggest that the improved calibration of SVGD in this setting may be associated with a trade-off in predictive precision. In contrast, DiffUQ maintains a more balanced performance across both accuracy and calibration metrics, as reflected by its consistently lower NLL even in the small-sample regime.

In terms of predictive accuracy, our method also consistently outperforms the baselines. Furthermore, under limited sample budgets, our approach maintains stable performance, whereas MFVI, SGLD, and SVGD all suffer from increased NLL. Notably, MFVI and MC dropout perform even worse than MAP, underscoring that restrictive structural assumptions on the posterior distribution (e.g., factorized Gaussian of MFVI) can be detrimental to the uncertainty-aware modeling.

In addition, the results obtained by our method exhibit overall smaller standard deviations compared with the baselines, which highlights the improved stability and robustness of our approach.

\subsubsection{Hyperparameter Sensitivity Analysis} This part is to demonstrate the robustness of DiffUQ.

\emph{Number of Samples $n$:} Results with different sample sizes are reported in \Cref{tab:hlt}. Even with as few as $n=4$ posterior samples, DiffUQ maintains comparable performance, exhibiting only a slight increase in variance. This behavior should be interpreted from the perspective of sample efficiency rather than sample quantity. The generated samples are highly informative, reflecting the amortized nature of the proposed diffusion-based inference procedure. Specifically, extensive SDE simulations during training enable thorough exploration of the posterior support, with the resulting structural information, such as multimodality and posterior geometry, encoded into the learned drift network. As a consequence, inference-time samples are not obtained through unguided random exploration, but are instead explicitly guided toward high-probability regions of the posterior. This separation between posterior exploration and inference-time sampling allows DiffUQ to achieve accurate uncertainty estimation with a small number of forward predictions, even in the presence of complex posterior structures.

\begin{figure*}[ht]
\centering
\begin{subfigure}[t]{0.80\textwidth}
\centering
\includegraphics[width=\linewidth]{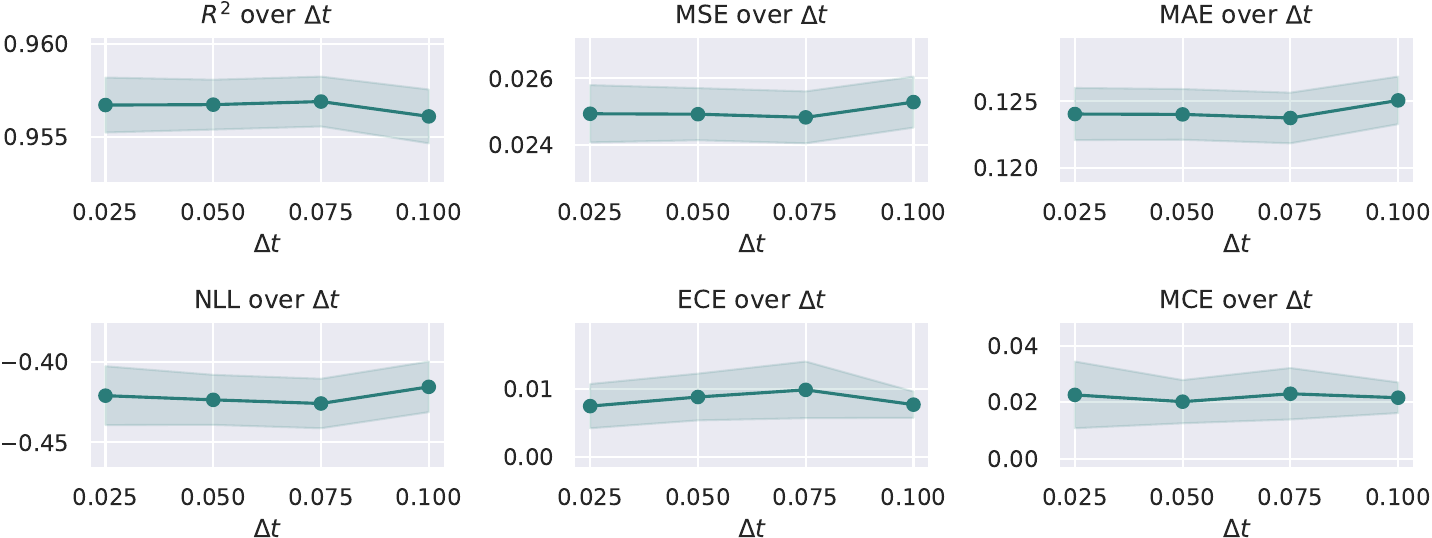}
\caption{Effect of different $\Delta t$ during training phase.}
\label{fig:overdt}
\end{subfigure}\vspace{20pt}
\begin{subfigure}[t]{0.80\textwidth}
\centering
\includegraphics[width=\linewidth]{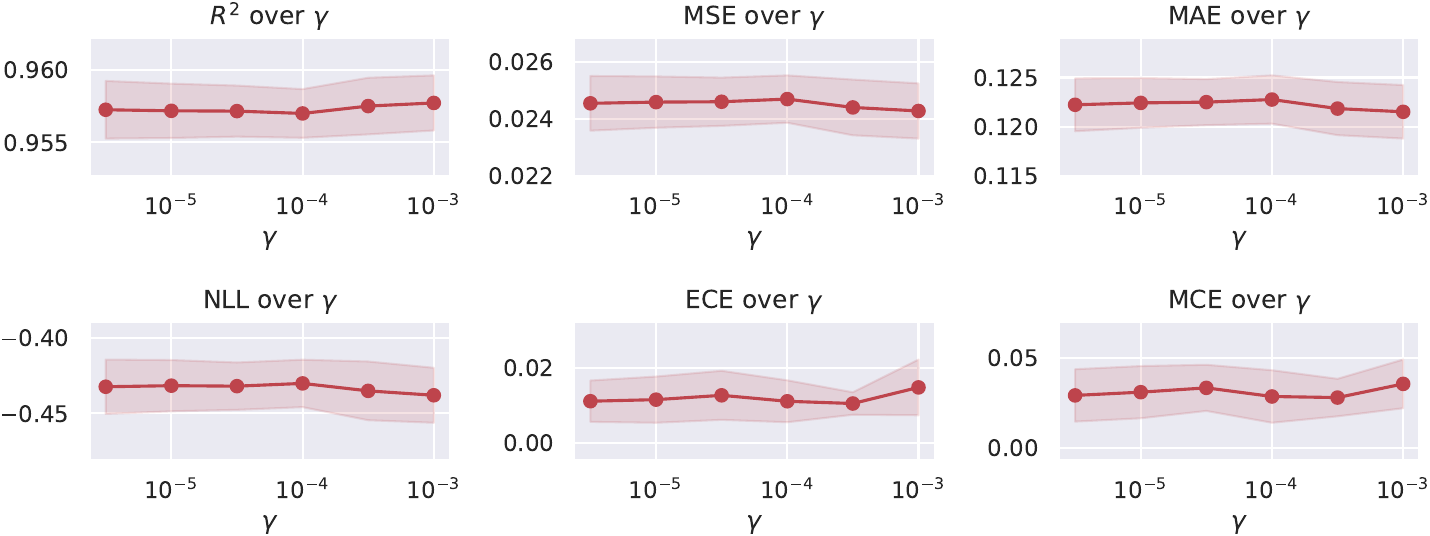}
\caption{Effect of different diffusion coefficient $\gamma$.}
\label{fig:overgamma}
\end{subfigure}
\caption{Sensitivity Analysis. The shaded area indicates $\pm 1$ standard deviation over 5 independent runs.}
\end{figure*}

\emph{Discretization Step Size $\Delta t$:} To assess the sensitivity of our method to hyperparameters, we first examined the effect of the discretization. Specifically, we varied the step size of EM discretization during the training phase, selecting $\Delta t \in \{0.025, 0.05, 0.075, 0.1\}$. The results in \Cref{fig:overdt} indicate that predictive accuracy is only slightly reduced when $\Delta t = 0.1$, while calibration metrics remain largely unaffected. This demonstrates that our method is robust to the choice of discretization step size.
From a theoretical perspective, this behavior is expected.
The optimization problem is defined over continuous-time trajectories, while $\Delta t$ only affects the numerical approximation of the SDE. Within a reasonable discretization regime, varying $\Delta t$ does not change the target objective but merely adjusts the resolution of the simulated paths.

\emph{Diffusion Coefficient $\gamma$:} We further conducted a sensitivity analysis with respect to the diffusion coefficient by varying its value across three orders of magnitude. The results in \Cref{fig:overgamma} consistently indicate that this hyperparameter has no meaningful influence on predictive accuracy or uncertainty quantification. Only when the diffusion coefficient was increased to $10^{-3}$ did we observe a minor degradation in calibration, while the other metrics remained unaffected. This robustness suggests that the method is relatively insensitive to the precise choice of the diffusion coefficient. $\gamma$ in our framework defines the volatility of the reference process. As requested by the theoretical analysis, these properties are not coincidental but stem from the stochastic control formulation. The optimization objective \eqref{eq:obj} drives a neural network to learn a drift function $u(t,\theta)$ that adaptively compensates for the noise level to reach the target posterior.
This creates a self-regulating mechanism where the learned control effort balances the diffusion intensity, ensuring consistent posterior sampling across orders of magnitude of $\gamma$.

\subsubsection{Training Dynamics}
\begin{figure}[ht]
\centering
\begin{subfigure}[t]{0.95\linewidth}
\centering
\includegraphics[width=\linewidth]{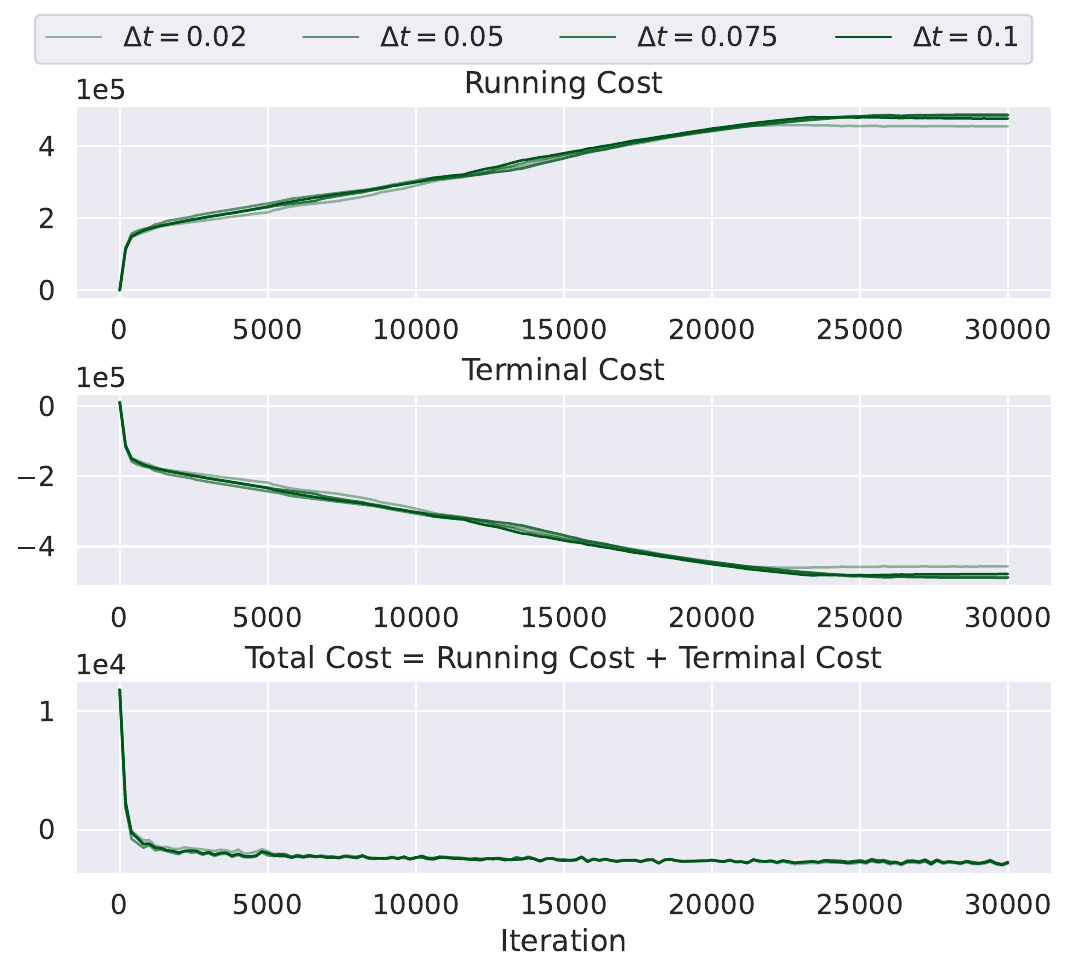}
\caption{Variation of $\Delta t$}
\label{fig:training_dt}
\end{subfigure}
% \hspace{30pt}
\begin{subfigure}[t]{0.95\linewidth}
\centering
\includegraphics[width=\linewidth]{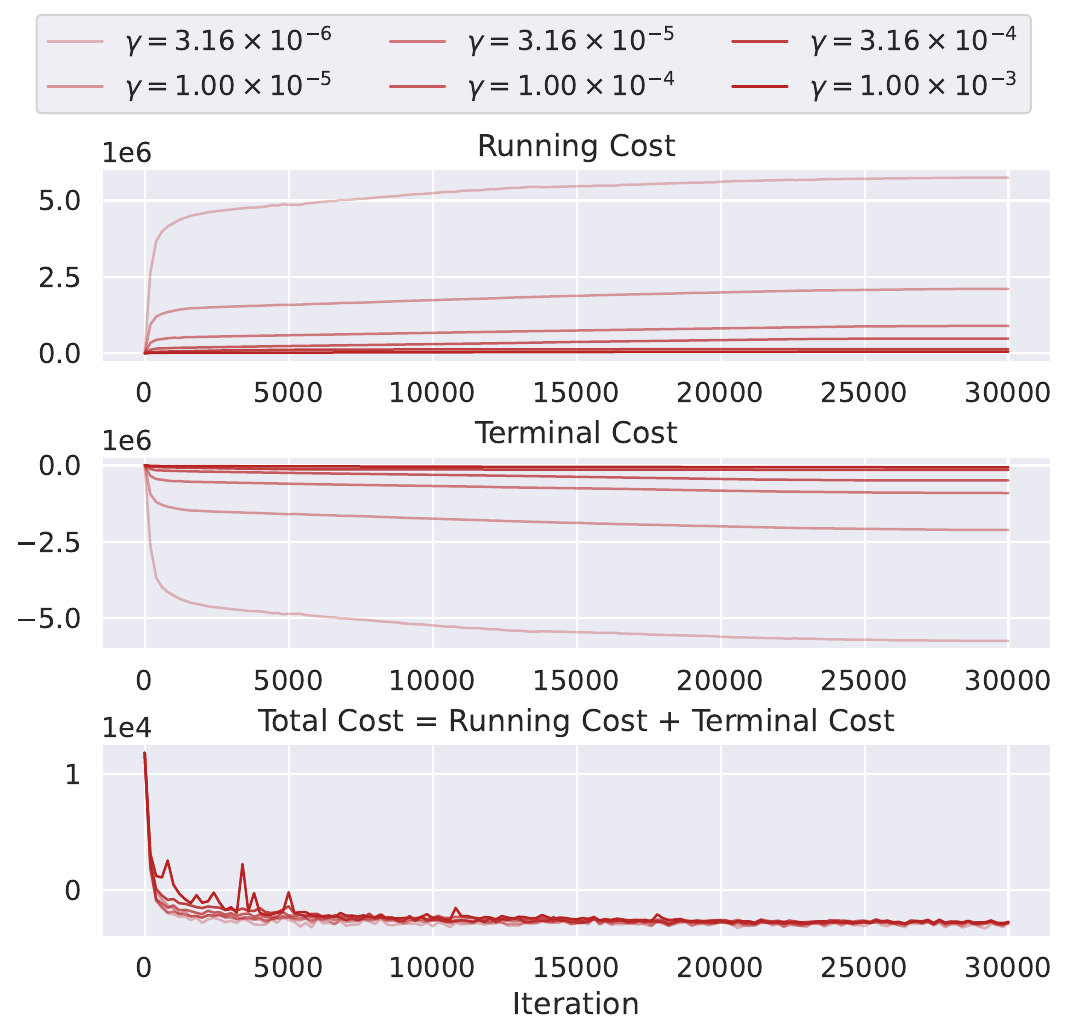}
\caption{Variation of $\gamma$}
\label{fig:training_gamma}
\end{subfigure}
\caption{Training Curves}
\label{fig:training}
\end{figure}

To provide a clearer understanding of DiffUQ's optimization behavior, we present the progression of its composite loss over multiple training runs (\Cref{fig:training}).
Accoring to (\ref{eq:obj}), the total loss compromises two competing terms: the running cost $\mathbb{E}\left[\int_0^1\frac{1}{2\gamma} \|u(t,\theta_t)\|^2 \mathrm{d}t\right]$, and the terminal cost $\mathbb{E}\left[-\log \frac{\pi_1(\theta_1)}{\mathcal{N}(\theta_1|0,\gamma \mathbf{I}_d)}\right]$.

Overall, the optimization process demonstrates stable and reproducible behavior across runs, with a gradual and monotonic balance established between the terminal and control objectives. As shown in \Cref{fig:training_gamma}, varying $\gamma$ mainly changes the relative scale of the running and terminal costs, while the overall convergence behavior remains smooth and consistent across runs. Similarly, for $\Delta t \in \{0.025, 0.05, 0.075, 0.1\}$, the optimization trajectories (\Cref{fig:training_dt}) exhibit stable and largely overlapping convergence patterns, with no noticeable oscillation or instability. These results indicate that DiffUQ maintains smooth and robust optimization dynamics across a broad range of discretization and diffusion hyperparameters.

The smooth optimization dynamics can be explained by the \emph{entropic regularization} intrinsically induced by the diffusion process \cite{leonard2013survey, chen2016relation, tzen2019theoretical}. Mathematically, the diffusion term acts as a smoothing operator on the probability density, convexifying the optimization landscape and preventing sharp minima. Furthermore, by relaxing the hard boundary constraints into a soft terminal penalty (Proposition 2), the framework avoids the numerical instabilities typical of constrained transport problems, resulting in a stable and monotonic convergence trajectory.

In addition, we notice that the total loss appears to stabilize after a few thousand steps; however, its decomposition indicates that training continues to evolve, revealing a trade-off relationship.
This behavior suggests that convergence should be evaluated by jointly monitoring both objectives, which together delineate an empirical Pareto front characterizing the balance between accuracy and regularization.

\section{Conclusion}
This work introduced DiffUQ, a new approach for uncertainty quantification in industrial models based on a diffusion sampler.
By leveraging elegant connections between sampling and optimal control, the method avoids strong assumptions on the posterior form that compromise fidelity, and can achieve robust representation of complex posterior landscapes through sufficient exploration enabled by SDE simulation during training.
These properties ensure reliable posterior sampling that directly translates into well-calibrated predictive uncertainty.
Experiments on toy distributions, a Raman-based PAA benchmark, and a process modeling task from a real-world ammonia synthesis imply that these theoretical advantages consistently yield improvements in both calibration and accuracy over existing baselines, without relying on post-hoc adjustments.
In addition, DiffUQ also exhibits robustness to hyperparameters and shows well-behaved training dynamics, which are particularly valuable in practical industrial applications where ease of deployment is of primary concern.
Overall, this work highlights the potential of diffusion samplers as a principled and scalable family of methods for advancing uncertainty quantification in complex industrial data-driven modeling.
Future research may further extend this framework by developing more efficient training strategies and exploring alternative diffusion formulations. Specifically, investigating other advanced solvers within the diffusion family, such as Denoising Diffusion Samplers (DDS) \cite{vargas2023denoising} or Adjoint Sampling (AS) \cite{havens2025adjoint}, could offer further potential to optimize the training dynamics and sampling efficiency for specific industrial scenarios.

\appendix
\subsection{On Drift Network Capacity}\label{apdx:capacity}
\subsubsection{Drift Network Capacity and Discretization Error}

\begin{figure}[h]
\centering
\begin{subfigure}{0.49\linewidth}
\centering
\includegraphics[width=\linewidth]{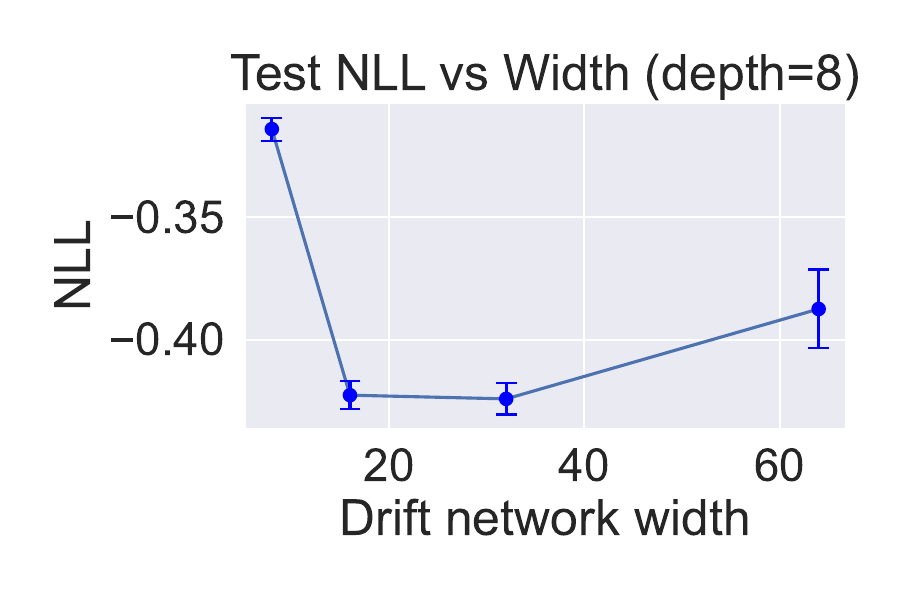}
\caption{NLL vs. Width}
\label{fig:width_nll}
\end{subfigure}
\begin{subfigure}{0.49\linewidth}
\centering
\includegraphics[width=\linewidth]{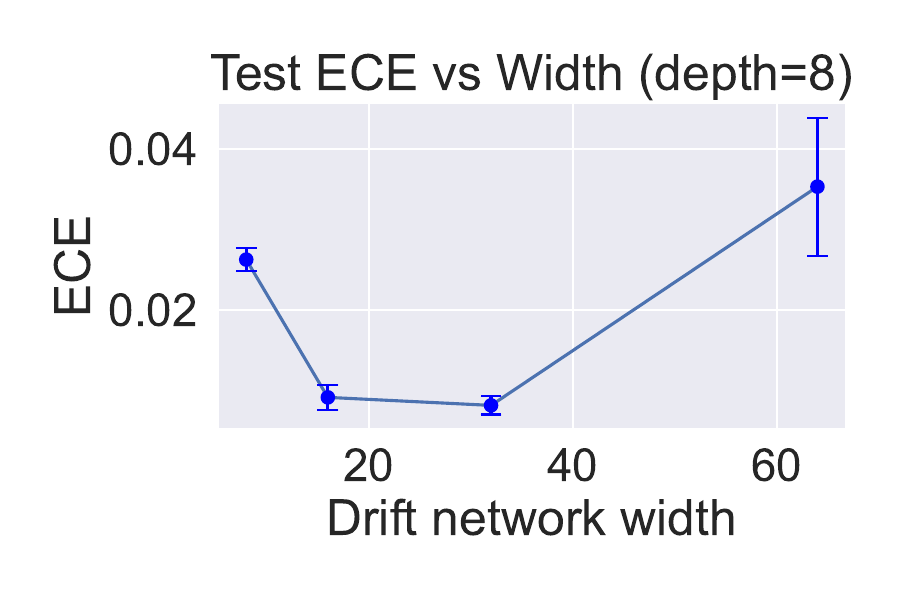}
\caption{ECE vs. Width}
\label{fig:width_ece}
\end{subfigure}
\begin{subfigure}{0.49\linewidth}
\centering
\includegraphics[width=\linewidth]{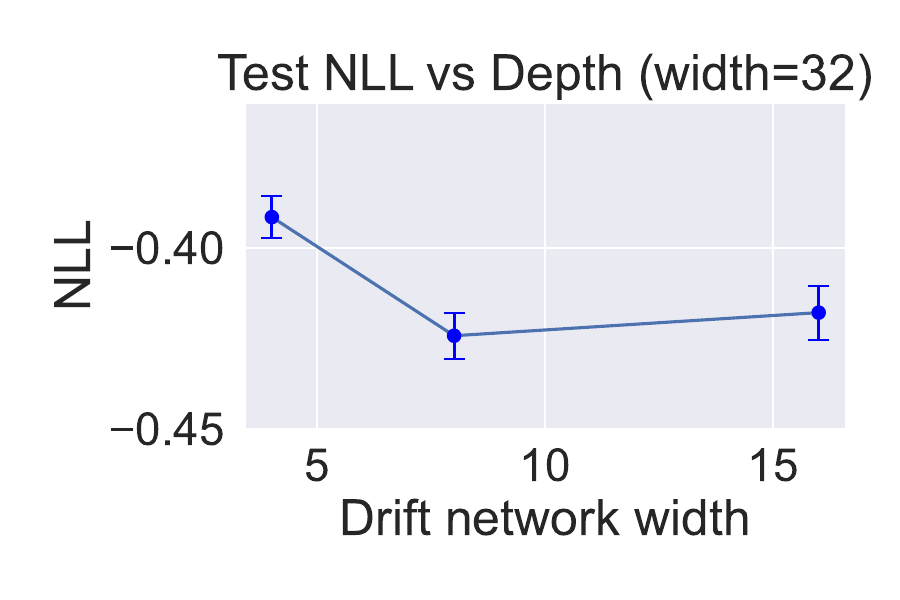}
\caption{NLL vs. Depth}
\label{fig:depth_nll}
\end{subfigure}
\begin{subfigure}{0.49\linewidth}
\centering
\includegraphics[width=\linewidth]{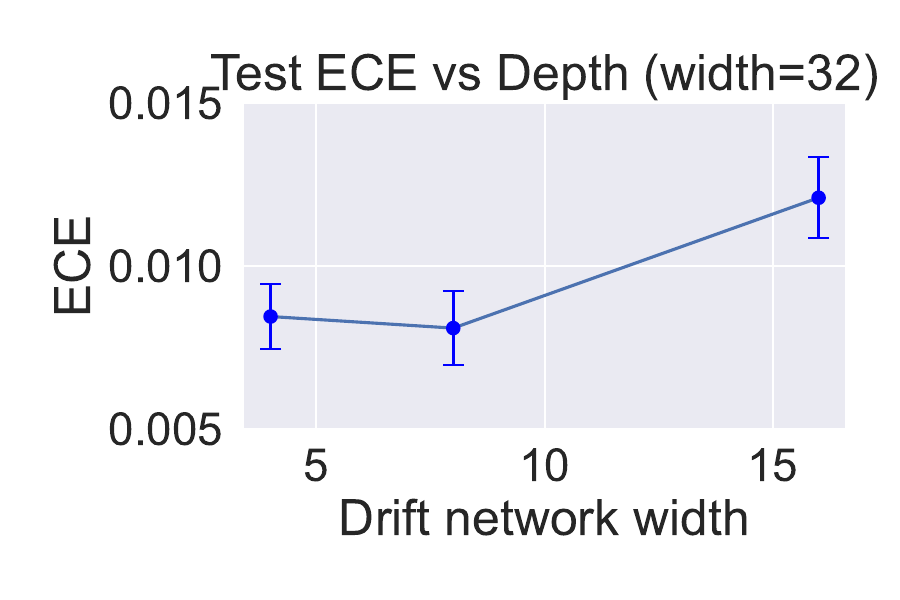}
\caption{ECE vs. Depth}
\label{fig:depth_ece}
\end{subfigure}
\caption{Impact of Drift Network Size}
\label{fig:network_capacity}
\end{figure}

From a theoretical perspective, existing results, e.g., Theorem~2 in \cite{zhang2021path}, ensure that a sufficiently expressive neural drift can approximate the target distribution in KL to arbitrary accuracy under mild regularity assumptions. However, these are \emph{existence} results and do not account for finite-step numerical discretization. In practice, the learned drift is simulated using a fixed-step Euler--Maruyama scheme. Under global Lipschitz and linear growth conditions, its strong error satisfies
\[
\mathbb{E}\Big[\sup_{s \le T} \|X_s - \hat X_s\|^2 \Big] \le C(L,T)\, h,
\]
where the constant $C(L,T)$ depends on the regularity of the drift via a Gr\"onwall argument (e.g., Theorem~10.2.2 in \cite{kloeden1992numerical}). While the convergence order is governed by the step size $h$, the prefactor is sensitive to the drift’s Jacobian magnitude. Increasing network capacity expands the hypothesis space and can lead to drifts with larger Jacobians, making fixed-step linear integration less accurate and effectively enlarging $C(L,T)$.

As a result, under a fixed discretization budget, larger networks do not necessarily improve sampling accuracy. Instead, there exists a practically relevant capacity regime in which the drift is sufficiently expressive, while further increasing network size mainly amplifies discretization error rather than improving posterior approximation.

This observation is supported by the ablation results on the CO soft sensor case study (\Cref{fig:network_capacity}). Empirically, we observe a non-monotonic dependence on network capacity: small networks suffer from insufficient expressivity, whereas overly large networks offer no further benefits and can even degrade both likelihood and calibration under a fixed step size, likely due to amplified discretization effects.

\subsubsection{Other drift network parameterization}
Exploring more suitable drift architectures and parameterizations remains an important direction for future work. \cite{zhang2021path} augment the time variable $t$ with Fourier features or sinusoidal embeddings to enhance temporal representation. However, our preliminary experiments did not show clear improvements.

\bibliographystyle{IEEEtran}
\bibliography{references}

\end{document}